\pdfoutput=1

\documentclass[11pt]{article}

\usepackage{EACL2023}

\usepackage{times}
\usepackage{latexsym}

\usepackage[T1]{fontenc}

\usepackage[utf8]{inputenc}

\usepackage{microtype}

\usepackage{inconsolata}
\usepackage{amsmath}
\usepackage{bm}
\usepackage{graphicx}
\usepackage{multirow}
\usepackage{colortbl}
\usepackage{amsfonts}
\usepackage{algpseudocode}
\usepackage{subfig}
\usepackage{arydshln}
\usepackage{booktabs}

\title{Plan-then-Seam: Towards Efficient Table-to-Text Generation}

\author{Liang Li$^{1, 2}$, Ruiying Geng$^{3}$, Chengyang Fang$^{1, 2}$, Bing Li$^{1}$, Can Ma${}^1$\thanks{$^{*}$Corresponding authors: Can Ma, Yongbin Li}, \\ \textbf{Binhua Li}$^{3}$ \textbf{,} \textbf{Yongbin Li}$^{3*}$ \\
$^1$Institute of Information Engineering, Chinese Academy of Sciences, Beijing, China \\
$^2$School of Cyber Security, University of Chinese Academy of Sciences, Beijing, China\\
$^3$DAMO Academy, Alibaba Group \\
\texttt{\{liliang, macan, fangchengyang\}@iie.ac.cn} \\
\texttt{\{ruiying.gry, binhua.lbh, shuide.lyb\}@alibaba-inc.com}
}
\begin{document}
\maketitle
\begin{abstract}
Table-to-text generation aims at automatically generating text to help people conveniently obtain salient information in tables. 
Recent works explicitly decompose the generation process into content planning and surface generation stages, employing two \textit{autoregressive} networks for them respectively.
However, they are computationally expensive due to the non-parallelizable nature of autoregressive decoding and the redundant parameters of two networks.
In this paper, we propose the first totally \textit{non-autoregressive} table-to-text model (Plan-then-Seam, PTS) that produces its outputs in parallel with one single network.
PTS firstly writes and calibrates one plan of the content to be generated with a novel \textit{rethinking} pointer predictor, and then takes the plan as the context for seaming to decode the description.
These two steps share parameters and perform iteratively to capture token inter-dependency while keeping parallel decoding.
Experiments on two public benchmarks show that PTS achieves $3.0\sim5.6$ times speedup for inference time, reducing 50\% parameters, while maintaining as least comparable performance against strong two-stage table-to-text competitors
\footnote{\url{https://github.com/liang8qi/Plan-then-Seam}}.
\end{abstract}

\section{Introduction}

Table-to-text generation \citep{DBLP:conf/emnlp/LebretGA16,DBLP:conf/emnlp/WisemanSR17} is a long-standing problem that aims to generate natural language descriptions of structured table data. A good table-to-text system can help end users better identify the informative elements and their relations from a table. Therefore, developing table-to-text systems is of tremendous value in a wide range of applications, such as biography generation \citep{DBLP:conf/emnlp/LebretGA16}, basketball news generation \citep{DBLP:conf/emnlp/WisemanSR17}, advertising text generation \citep{DBLP:conf/emnlp/ShaoHWXZ19}, and table-based question answering \citep{DBLP:conf/emnlp/YuZELXPLTSLJYSC19,Fu_2020_arxiv_survey_on_question}.

\begin{figure}[t]
\centering
\includegraphics[width=0.95\columnwidth]{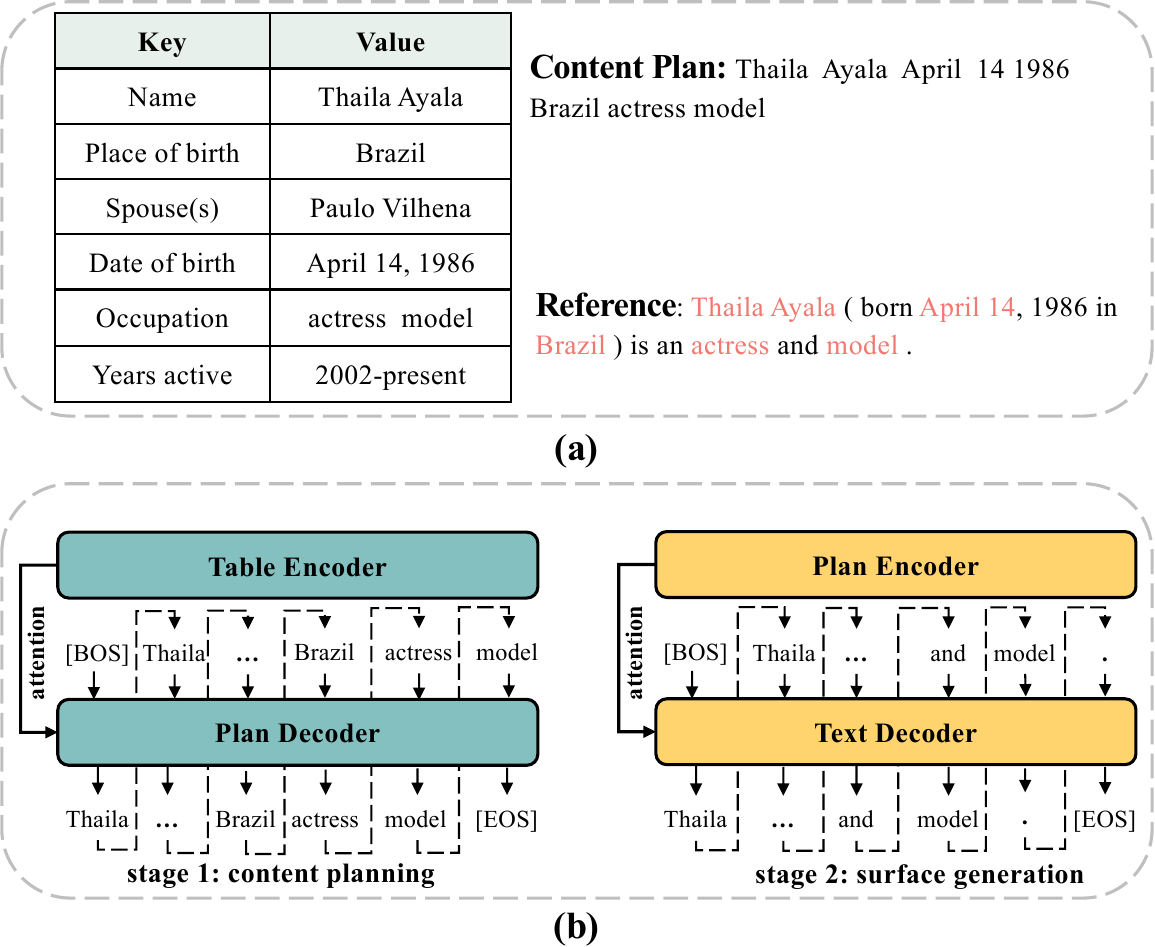}
\caption{(a): example of table-to-text generation from WikiBio. Tokens from the content planning are colored in red. (b): two-stage model, which disentangles table-to-text generation into two stages: content planning and surface generation.}
\label{fig:introduction}
\end{figure}
Recently, neural network-based approaches have made significant progress in this field.
The modern neural models for table-to-text generation can be roughly categorized into one-stage models and two-stage models.
One-stage models generate natural language descriptions directly from the table by simply relying on representation learning to generate well-organized fluent descriptions.
Along this line, some studies propose to modify the model architectures to effectively learn from structured table~\citep{DBLP:conf/aaai/LiuWSCS18,DBLP:conf/emnlp/GongFQL19}, while some other works introduce auxiliary tasks to help the encoder capture a more accurate semantic representation \citep{DBLP:conf/aaai/LiuLXMCS19,DBLP:conf/acl/LiMYH20}.
The major drawback of one-stage models is the lack of interpretability and controllability, making the models prone to suffer from unfaithful hallucinations \citep{DBLP:conf/emnlp/WisemanSR17}.

To alleviate the aforementioned shortcoming of one-stage models, some researchers propose a two-stage paradigm for table-to-text generation 
\citep{DBLP:conf/aaai/Puduppully0L19,su2021plan}, which explicitly decomposes the whole generation process into two separate stages: content planning and surface generation (illustrated in Figure \ref{fig:introduction}). The content planning model generates an intermediate sequence that specifies the tokens to be verbalized. The generated plan provides some interpretability and controllability, thus can potentially reduce the risk of hallucinations \citep{DBLP:conf/aaai/Puduppully0L19}. The surface generation model then completes the description based on the intermediate plan.

Although two-stage models have some superiority over one-stage models, they are often computationally intensive. The cause of the high computational cost is two-fold.
Firstly, most two-stage models use autoregressive (AR) decoders, which is quite time-consuming due to their non-parallelizable nature, especially for long sequences \citep{gu2018non}.
Secondly, the two-stage systems often consist of two different models, which usually double the parameter scale (see Section~\ref{sec:main_results}). The increased parameter scale may introduce more computation overhead and thus slow down the inference speed.
These disadvantages limit the deployment of current neural table-to-text systems in practical applications. Recently, non-autoregressive (NAR) generation has attracted much attention because it can significantly accelerate inference speed for text generation~\citep{gu2018non,DBLP:conf/icml/SternCKU19,DBLP:conf/acl/QianZBWQ0YL20}. However, as demonstrated in our preliminary experiments (see also Section~\ref{sec:main_results}), applying the NAR models directly to table-to-text generation may suffer from lower generation quality because NAR models do not explicitly model the content planning procedure, which can provide good initial input for NAR decoder.

In this work, we propose to reduce the computational cost of two-stage models through a unified NAR framework, which is called {\em Plan-then-Seam} (\textproc{PTS}). PTS is a iterative NAR table-to-text model. Specifically, \textproc{PTS} first generates the content plan in the first iteration.
Then it fills in the other surface tokens in subsequent multiple iterations to seam the intermediate plan tokens.
Note that PTS conducts the content planning and surface generation tasks in a single model, thus the model size is smaller than previous two-stage models. Moreover, since PTS is a NAR model, it is more efficient than the AR counterparts.
Given that fully NAR content planning may ignore the dependency between planned tokens, we introduce a rethinking pointer predictor, which can better calibrate the planning conditioned on the generated ones.
Our contributions can be summarized as follows:
\begin{itemize}
    \setlength{\itemsep}{0pt}
    \setlength{\parsep}{0pt}
    \setlength{\parskip}{0pt}
    \item We are the first work concerning the computational cost (parameter and inference efficiency) problem in table-to-text. Contrastly, previous works only focus on how to improve the model performance. We hope this can raise more attention to the computational cost problem in table-to-text.
    
    \item Regarding methodology, we present the first totally NAR \footnote{This means both content planning and surface generation are non-autoregressive.} framework for table-to-text generation, achieving a desired quality–efficiency trade-off. Further, we introduce a rethinking mechanism to improve the NAR planning capability of the model. 
    We demonstrate that initializing the decoder with a good content plan is the key to improving the NAR model.
    \item Experiments show that, compared with previous strong two-stage competitors, our method can achieve a $3.0\sim5.6\times$ speedup with only 50\% model parameters without degrading the generation quality.
\end{itemize}

\section{Related Work}
Table-to-text generation has long aroused interest in the Natural Language Generation (NLG) community \citep{DBLP:conf/acl/Kukich83,DBLP:journals/nle/ReiterD97}. Recently, neural models have been the mainstream for this task and made impressive progress. Models for this task can be mainly categorized into two types: one-stage models and two-stage models. One-stage models generate text directly from structured data through a neural encoder-decoder architecture \citep{DBLP:conf/nips/SutskeverVL14}. They simply rely on representation learning to improve the generation. \citet{DBLP:conf/aaai/LiuWSCS18} propose a structure-aware seq2seq architecture, which incorporates the filed information as the additional inputs to the table encoder. Some works design hierarchical table encoder which model table's representation from the row and column levels \citep{DBLP:conf/emnlp/GongFQL19}. \citet{DBLP:conf/aaai/LiuLXMCS19,DBLP:conf/acl/LiMYH20} introduce auxiliary supervision tasks to help the encoder capture a more accurate semantic representation of the tables. However, one-stage methods are prone to produce unfaithful hallucinations and uncontrollable generation \citep{DBLP:conf/emnlp/WisemanSR17}.

As the improvement, neural two-stage models \citep{DBLP:conf/acl/MaYLLZS19,DBLP:conf/aaai/Puduppully0L19,DBLP:conf/naacl/MoryossefGD19,DBLP:journals/tacl/PuduppullyL21,su2021plan} decompose the table-to-text generation into content planning and surface generation stages. In general, content planning is implemented by Pointer Networks \citep{DBLP:conf/nips/VinyalsFJ15}. The explicit content planning mechanism not only decomposes the complex table-to-text generation into two easier tasks but also makes the generation process more interpretable and controllable by generating an intermediate representation. However, the hallucination problem persists in the surface generation stage as it is autoregressive (AR). To address this issue, SANA \citep{DBLP:conf/acl/WangLYZZZY21} proposes an edit-based non-autoregressive (NAR) surface generation model that generates texts through iterative insertion and deletion operations while maintaining an AR planning stage. 
Existing two-stage methods solely pay attention to improving the generation quality while ignoring its efficiency. Compared with one-stage models, two-stage methods double the number of parameters. Additionally, the AR generation is slow at inference time. These problems hinder the practical deployment of current neural table-to-text models.

\begin{figure*}[t]
\centering
\includegraphics[width=2.0\columnwidth]{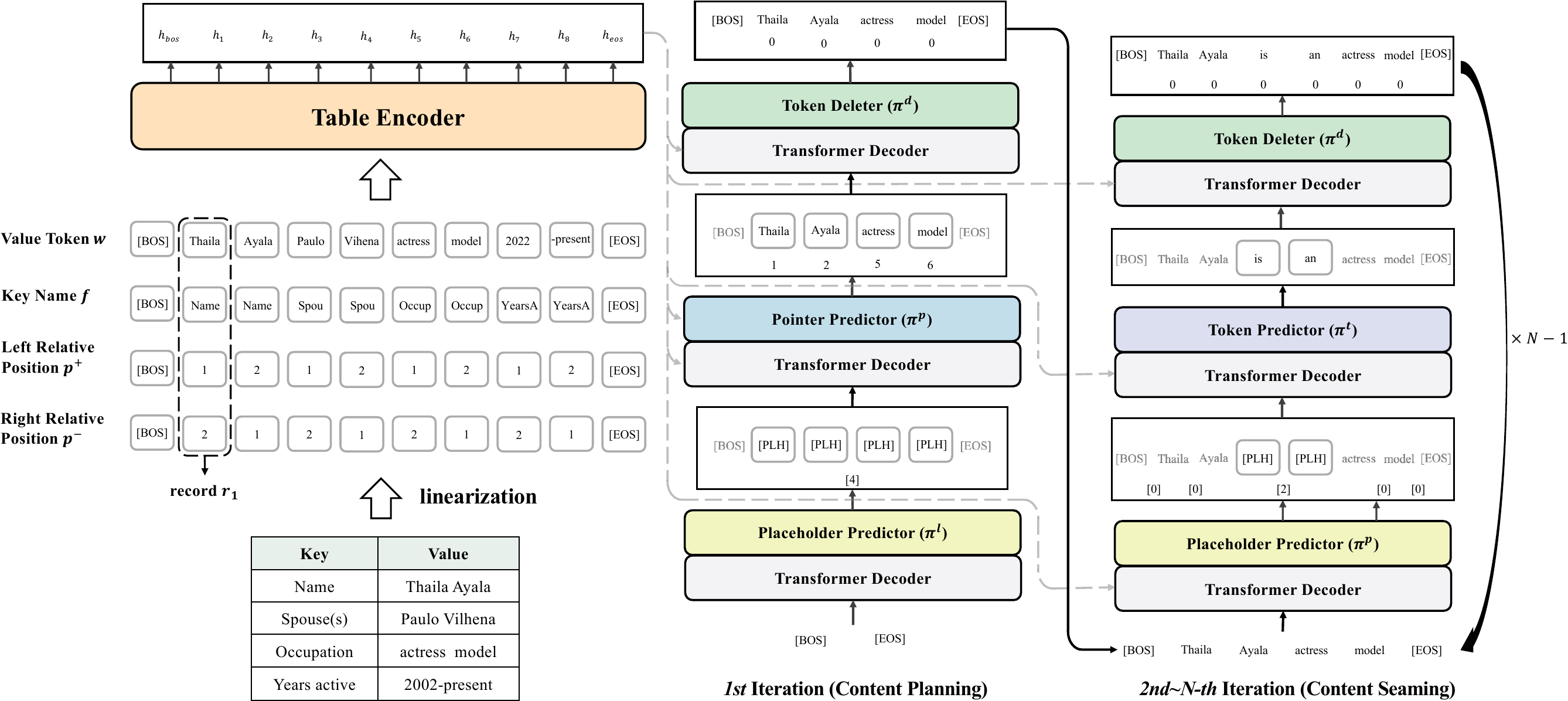}
\caption{An overview of our Plan-then-Seam non-autoregressive table-to-text model. The modules with the same color share the parameters. The left part contains an example of how to linearize a table, where  \texttt{Spou}, \texttt{Occup}, and \texttt{YearsA} denote \texttt{Spouse(s)}, \texttt{Occupation}, and \texttt{Years active}, respectively.
}
\label{fig:framwork}
\end{figure*}

\section{Methodology}
Given a region of a table as input, table-to-text generation is to produce
a natural language description $Y = \{y_1, ..., y_n\}$ to describe the selected table
region.
This paper proposes the first totally non-autoregressive table-to-text model, {\em Plan-then-Seam} (\textproc{PTS}).
As depicted in Figure \ref{fig:framwork}, \textproc{PTS} consists of three major components, a table encoder, a non-autoregressive content planning decoder (NAR-P), and a non-autoregressive seaming decoder (NAR-S), which collaborate to generate a description for a source table in an iterative manner. 
At the first iteration, NAR-P generates content planning sequence in a fully non-autoregressive manner by conditioning on the source table. 
At the subsequent iterations, NAR-S seams the content planning tokens by inserting connective tokens between them to generate a fluent description. 
Next, we will describe the proposed \textproc{PTS} in detail.
 
\subsection{Table Encoder}
\label{subsection:table_encoder}
As shown on the left of Figure \ref{fig:framwork}, the source table is a collection of key-value pairs in which each value may contain several tokens. Following \citet{DBLP:conf/emnlp/LebretGA16}, we first linearize the source table by flattening all its values to a record sequence $T = \{r_1, r_2, ..., r_K\}$.
Each record $r_i$ is represented as a $4$-tuple $(w_i, k_i, p^+_i, p^-_i)$, where $w_i$ is the value token, $k_i$ is its key name. 
$p^+_i$ and $p^-_i$ are the relative positions of $w_i$, where $p^+_i$ counts from the beginning and $p^-_i$ counts from the end of the sentence. For example, the key-value pair <\verb|Name|, \verb|Thaila Ayala|> is represented as two records: (\verb|Thaila|, \verb|Name|, 1, 2) and (\verb|Ayala|, \verb|Name|, 2, 1). 
We adopt four trainable embedding matrices to convert each record represented by $(w_i, k_i, p^+_i, p^-_i)$ into dense vectors $e_{w_i}$, $e_{k_i}$, $e_{p^+_i}$, and $e_{p^-_i}$.
We concatenate these embeddings and use a linear projection to map the four vectors into $\mathbf{e}_i$, which serves as the initial representation of the corresponding record $r_i$:
\begin{align}
    \mathbf{e}_i = {\rm ReLU}(\mathbf{W}_e[e_{w_i}; e_{k_i}; e_{p^+_i}; e_{p^-_i}] + \mathbf{b}_e),
\end{align}
where $\mathbf{W}_e$ and $\mathbf{b}_e$ are trainable parameters. $[\cdot ; \cdot]$ denotes the vector concatenation operation. Finally, we transform $\{\mathbf{e}_1, \mathbf{e}_2, ..., \mathbf{e}_K\}$ into contextual sequence representation $\mathbf{H}^e=\{\mathbf{h}^e_1, \mathbf{h}^e_2, ..., \mathbf{h}^e_K\}$ with the Transformer encoder \citep{DBLP:conf/nips/VaswaniSPUJGKP17}. 

\subsection{Non-autoregressive Content Planning Decoder}
We utilize the Transformer decoder layer \citep{DBLP:conf/nips/VaswaniSPUJGKP17} as the basic building block of the content planning component. We also remove the causal mask in self-attention modules to realize parallel generation. 
As shown in Figure \ref{fig:framwork}, given the initial decoder input $y^0 = $\verb|[BOS][EOS]|, non-autoregressive planning decoder (NAR-P) aims to generate the planned sequence (e.g., $y^p = $ \verb|Thaila Ayalia actress model|).  
$y^p$ specifies the records that are to be verbalized ({\em what to say}) in the description and the order in which they are described. To this end, NAR-P consists of three major components: a placeholder predictor $\pi^{l}$, a pointer predictor $\pi^p$, and a token deleter $\pi^d$.
These components work in a serial fashion. 
First, the placeholder predictor $\pi^{l}$ determines the number of plan tokens to be inserted: 
\begin{align}
    \pi^l(l|y^0, T) &= {\rm Softmax}(\mathbf{W}_l[\mathbf{h}^{d_1}_{0};\mathbf{h}^{d_1}_{1}]),
\end{align}
where $\mathbf{h}^{d_1}_{0}$ and $\mathbf{h}^{d_1}_{1}$ are respectively the decoder states of two symbol tokens in $y^0$, $\mathbf{W}_{l}\in \mathbb{R}^{L\times 2d}$ is the projection matrix and $L$ is the pre-defined maximal placeholder number. 
$ \pi^l(l) \in \mathbb{R}^{L}$ denotes the the probability distribution of possible placeholder numbers, and we choose the one $l$ with the highest probability. 
We insert $l$ placeholders \verb|[PLH]| between \verb|[BOS]| and \verb|[EOS]| to obtain the placeholder sequence $y^{l}$.

Then, we need to replace each \verb|[PLH]| in $y^{l}$ with an actual token. 
$y^{l}$ is firstly passed into the Transformer decoder layer to generate the decoder state $\mathbf{h}^{d_2}_{i}$ of each placeholder.
As all plan tokens are from the source table, we then introduce a pointer predictor $\pi^p$ that selects tokens from $T$ to reduce hallucinations. Specifically, for the $i$-th placeholder, we calculate the confidence scores $\alpha^i_j$ of copying the $j$-th record in $T$ as a plan token by:
\begin{equation}
\label{equaltion:naive_pointer_predictor}
    \pi^p(\alpha^i_j|y^l, T)={\rm Softmax}(
        \mathbf{W}_{p} [\mathbf{h}^{d_2}_{i};\mathbf{h}^e_j]
        ).
\end{equation}
Then we replace each placeholder with the most possible record to get $y^{r}$. 

Last, considering the predictor $\pi^p$ may copy incorrect or repetitive tokens, we build a token deleter $\pi^{d}$ to remove these false plan tokens. 
For the $i$-th token in $y^r$, $\pi^d$ is employed to decide whether it is required to be deleted or not:
\begin{align}
\pi^d(d_i|y^r, T) = {\rm Softmax}(\mathbf{W}_{d} \mathbf{h}^{d_3}_{i}),
\end{align}
where $\mathbf{h}^{d_3}_{i}$ is the representation generated by transformer decoder. $\pi^d(d_i) \in [0, 1]$ is the predicted probability of the deletion operation. The token with $\pi^d(d_i = 1) > 0.5$ is deleted, which yields the final content plan $y^p$.

\begin{figure}
    \centering
    \includegraphics[width=0.75\columnwidth]{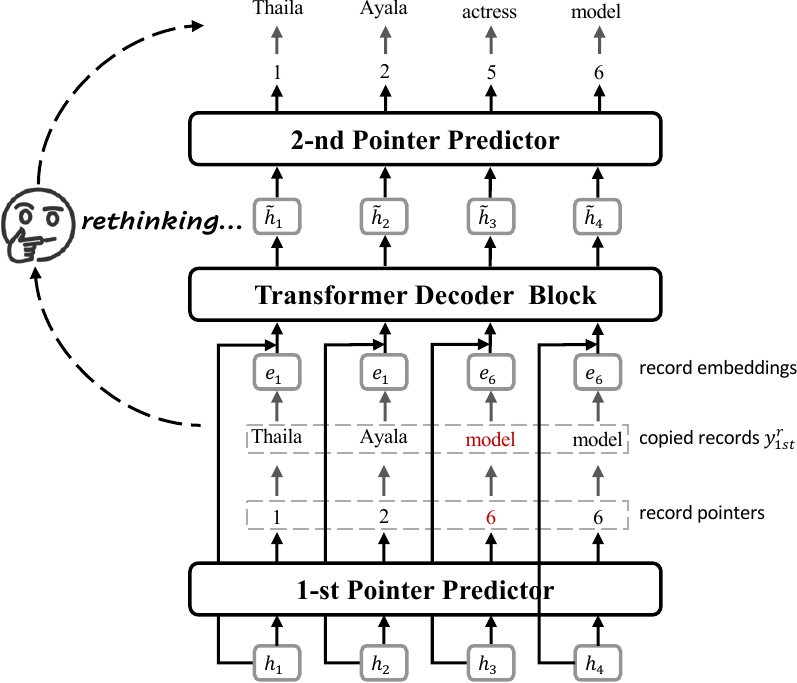}
    \caption{Architecture of the proposed {rethinking} pointer predictor. Our basic idea is to augment the non-autoregressive predictor with inter-token dependencies.
    }
    \label{fig:prediction_aware_pointer}
\end{figure}
\paragraph{Rethinking Pointer Predictor} 
Although non-autoregressive models can accelerate the generation process, they are based on the assumption that the generated tokens are conditionally independent with each other~\citep{gu2018non}.
As a result, the pointer predictor may suffer from  incoherence or repetition~\citep{DBLP:conf/acl/QianZBWQ0YL20}. 
We believe that the pointer predictor would produce a better plan by partially observing generated plan tokens. Motivated by this intuition, we adapt the naive pointer predictor and propose its variant \textit{rethinking} pointer predictor. As illustrated in Figure \ref{fig:prediction_aware_pointer}, the rethinking pointer predictor first employs a navie pointer predictor to generate a primary record plan  $y^r_{1st} = \{r_1, r_2, r_6, r_6\}$
, then calibrate it with another pointer predictor.
Particularly, for each record $r_i$ in $y^r_{1st}$, 
we concatenate its embedding $e_{i}$ with the transformer hidden state $h_i$ and further process it by a linear layer $\hat{h}_i = W_f[h_i;e_{i}]$. $\hat{h}_i$ is fed into a transformer decoder block (removing causal attention) to rebuild representation for the token at $i$-th position in $y^r_{1st}$ by conditioning on the source table $T$ and the generated plan $y^r_{1st}$. By looking at the tokens in other positions, the $2$-nd pointer predictor can determine if the token at the $i$-th position is incorrect or repeated with others.
The \textit{rethinking} process helps the model adjust the incorrect tokens in $y^r_{1st}$ to generate a precise plan. 
Moreover, to 
improve the confidence of $y^r_{1st}$, both pointer predictors are supervised by the ground truth plan when training.

\subsection{Non-autoregressive Content Seaming Decoder}
After obtaining the content plan $y^p$, at the seaming stage, the non-autoregressive seaming decoder (NAR-S) constructs a fluent description $y$ by iteratively inserting the connective tokens into $y^p$. 
Specifically, we employ $y^p$ as the initial input of NAR-S. In each iteration, 
similar to NAR-P, NAR-S also firstly predicts the number of placeholders should be inserted into at every consecutive position pairs in the generated sentence of the previous iteration, then replace placeholders with actual tokens, and finally delete abundant tokens.
NAR-S share parameters in placeholder predictor $\pi^l$ and token deleter $\pi^d$ with NAR-P. 
The difference is that NAR-S replaces the pointer predictor $\pi^p$ with a token predictor $\pi^t$ that replaces the placeholders with actual tokens from the predefined vocabulary rather than copies from the table.
For example, as shown on the right of Figure \ref{fig:framwork}, given the content plan \verb|Thaila Ayalia actress model|, NAR-S inserts two tokens,  \verb|is| and \verb|an|, between \verb|Ayalia| and \verb|actress| in the first iteration.
\subsection{Training}
We joint train the content planning and seaming tasks and the final learning objective is:
\begin{align}
    \mathcal{L} &= \lambda \mathcal{L}_{plan} + \mathcal{L}_{seam},
\end{align}
where $\lambda$ is the hyper parameter. Next, we describe them in detail.

The content planning learning objective consists of three sub-goals: $\mathcal{L}_{plan} = \mathcal{L}^p_{l} + \mathcal{L}^p_{p} + \mathcal{L}^p_{d}$. Given the source table $T$, the ground truth content plan $y^{p^*}$ as well as 
its pointers $y^{idx} = \{y^{idx}_i\}^{|y^{p^*}|}_{i=1}$ with each entry pointing to an input record in $T$,
the placeholder predictor learning objective $\mathcal{L}^p_{l}$ is computed as follows:
\begin{align}
    \mathcal{L}^p_{l} &= - \log \pi^l(l^{*}_0|y^0, T),
\end{align}
where $l^{*}_0$ is the length of ground truth plan $y^{p^*}$. 
And then, we replace all the tokens in $y^{p^*}$ with \verb|[PLH]| to get $y^{l}$, 
which is utilized to train the pointer predictor:
\begin{align}
    \mathcal{L}^p_{p} &= - \sum^{|y^{p^*}|}_{i = 1} \log \pi^p(\alpha^i_{y^{idx}_i}|y^l, T).
\end{align}
To train the deletion predictor, we apply the pointer predictor $\pi^p$ to $y^{l}$ to yeild $y^{r}$. The loss for deletion predictor is calculated as:
\begin{align}
    \mathcal{L}^p_{d} &= - \sum^{|y^r|}_{i = 1} \log \pi^d(d_i|y^r, T),
\end{align}
where $d_i$ is the golden deletion operation at the $i$-th position, and is set as 1 if $y^{r}_i$ is same with $y^{p^*}_i$, otherwise 0. 

The seaming loss also consists of three parts: $\mathcal{L}_{seam} = \mathcal{L}^s_{l} + \mathcal{L}^s_{p} + \mathcal{L}^s_{d}$. 
Its training process is very similar to content planning task. The biggest difference between them is the initial input.
Given a source table, a plan and a reference ($T$, $y^{p^*}$, $y^*$), we follow previous works \citep{DBLP:conf/nips/GuWZ19,DBLP:conf/acl/WangLYZZZY21} to construct an intermediate sequence $y^m$ as the initial input to NAR-S. Specially, we first calculate the longest common subsequence $\hat{y}$ between $y^{p^*}$ and $y^*$. And then we apply random deletion on $y^*$ except the part of $\hat{y}$ to obtain $y^m$. Last, the three subgoals are calculated as following: 
\begin{align}
    \mathcal{L}^s_{l} &= - \sum^{|y^m|}_{i = 1} \log \pi^l({l}^*_i|y^m, T),
     \\
    \mathcal{L}^s_{p} &= - \sum^{|y^*|}_{i = 1} \log \pi^t(y^*_i|y^l, T), \\
   \mathcal{L}^s_{d} &= - \sum^{|y^*|}_{i = 1} \log \pi^d(d_i|y^t, T),
\end{align}
where $l^*_i$ is the number of placeholder that should be inserted between $y^m_i$ and $y^m_{i+1}$. $l^*_i$ is obtained by calculating the Levenshtein distance \citep{levenshtein1966binary} between $y^m$ and $y^*$. $y^t$ is yielded by applying $\pi^t$ on $y^l$.

\subsection{Inference}
As mentioned above, \textproc{PTS} is an iterative NAR model. Different from the previous iterative NAR model, at the first iteration, \textproc{PTS} first utilizes NAR-P to generate the content plan in a fully NAR manner, where NAR-P alternately performs placeholder prediction, pointer prediction, and deletion operation. In the subsequent iterations, it uses the generated plan as the initial decoder input for NAR-S, which iteratively fills in the other surface tokens between the content planning tokens. We stop the seaming process when the current text does not change or the predefined maximum iteration has been reached.

\section{Experiment}
\subsection{Datasets and Evaluation Metrics}
Following \citeauthor{DBLP:conf/acl/WangLYZZZY21}, we conduct experiments on two datasets, WikiBio~\citep{lebret2016neural} and WikiPerson~\citep{wang2018describing}.
Both datasets are designed to generate descriptions from a Wikipedia table. Specifically, WikiBio aims to generate the first sentence of a biograph. The average length of the description is 26.1 tokens. Different from Wikibio, the reference of WikiPerson contains multiple sentences to cover as many factors in the source table as possible. 
The average length of the description in WikiPerson is 70.6. We use the official training, development, and test splits for both datasets, which are 582,657/72,831/72,831 in WikiBio and 250,186/30,487/29,982 in WikiPerson.
We use these two datasets for two considerations. First, this paper focuses on the inference speed and generation quality of models with similar frameworks. The similar input structures allow us to use the same encoder architecture and prevent us from designing an additional one. 
Second, the different output length distributions of the two datasets facilitate us to compare the models' performance and efficiency.

We use BLEU \citep{papineni2002bleu} and ROUGE-L to evaluate the fluency, and PARENT \citep{DBLP:conf/acl/DhingraFPCDC19} to examine the faithfulness. 
We also employ inference latency to evaluate the inference speed of the involved approaches. Specifically, Latency is the average time to run an epoch on the test dataset while the batch size is set to $32$ with one NVIDIA Tesla V100 GPU.

\subsection{Baselines}
To rule out the effect of model architecture on the inference speed, we only compare our method to some representative baselines built on the Transformer~\citep{DBLP:conf/nips/VaswaniSPUJGKP17} model:

\begin{itemize}
    \setlength{\itemsep}{0pt}
    \setlength{\parsep}{0pt}
    \setlength{\parskip}{0pt}
    \item \textproc{TableTransformer} is a transformer-based model that replaces the naive transformer encoder with the table encoder.
    \item \textproc{LevT} \citep{DBLP:conf/nips/GuWZ19} is an iterative NAR model. In the first iteration, the decoder input is initialized by ``\verb|[BOS][EOS]|''.
    \item \textproc{Content-Plan} \citep{DBLP:conf/aaai/Puduppully0L19} is a representative two-stage method that firstly uses a pointer network to generate the content plan and then uses a pointer generator to complete the remaining text. To make a fair comparison, we reimplement it using Transformer. See Appendix~\ref{sec:appendix_experimental_setting} for more details.
    \item \textproc{SANA} \citep{DBLP:conf/acl/WangLYZZZY21} is also a two-stage method. The major difference between \textproc{SANA} and \textproc{Content-Plan} lies in that \textproc{SANA} uses a \textproc{LevT} for surface token generation. Additionally, \textproc{SANA} integrates hard constraints by forbidding the \textproc{LevT} from deleting planned tokens.
\end{itemize}

\begin{table*}[t]
    \centering
    \subfloat[Results on WikiBio]{
    \resizebox{\textwidth}{!}{
    \begin{tabular}{l l l l r r r}
    \hline
     \rowcolor[RGB]{237,237,237} \bf Models & \bf BLEU & \bf ROUGE-L & \bf PARENT (P / R / F1) & \bf \#Param & \bf Latency$\downarrow$ & $\mathbf{I_{DEC}}$ $\downarrow$  \\
     \hline
    \multicolumn{7}{c}{\em One-Stage Systems} \\
    \hline
 
    \textproc{TableTransformer} & {\bf 44.32}$_{\pm 0.32}$ & {\bf 66.75}$_{\pm 0.36}$ & {\bf 74.09}$_{\pm 0.32}$ / {\bf 42.41}$_{\pm 0.18}$ / {\bf 51.76}$_{\pm 0.31}$ & 76 & 680 & 22.10 \\
    \textproc{LevT} & 43.05$_{\pm 0.21}$ & 65.61$_{\pm 0.31}$ & {72.22}$_{\pm 0.16}$ / 37.62$_{\pm 0.14}$ / 47.14$_{\pm 0.11}$ & \bf 74 & \bf 223 & \bf 2.48 \\

    \hline
    \multicolumn{7}{c}{\em Two-Stage Systems} \\
    \hline

    \textproc{Content-Plan} & 43.44$_{\pm 0.00}$ & 66.21$_{\pm 0.31}$ & 74.55$_{\pm 0.29}$ / 43.45$_{\pm 0.30}$ / 52.38$_{\pm 0.11}$ & 150 & 1,381 & 30.47 \\
    \textproc{SANA} $\dag$ & \bf 45.78 & - & 76.93\ \quad\quad/ \textbf{46.01}\ \quad\quad/ {\bf 55.42} & - & - & - \\
    \quad w/o hard constraints $\dag$ & 45.31 & - & 76.32\ \quad\quad/ 45.26\ \quad\quad/ 54.64 & - & - & - \\
    \textproc{SANA} & 45.50$_{\pm 0.13}$ & 67.98$_{\pm 0.15}$ & 77.01$_{\pm 0.25}$ / 45.52$_{\pm 0.08}$ / 55.16$_{\pm 0.10}$ & 148 & 756 & 11.02 \\
    \quad w/o hard constraints & 44.94$_{\pm 0.16}$ & 67.72$_{\pm 0.16}$ & 76.89$_{\pm 0.26}$ / 44.70$_{\pm 0.26}$ / 54.68$_{\pm 0.59}$ & 148 & 761 & 11.03 \\
    \hdashline
    \textproc{Ours} & 45.65$_{\pm 0.13}$ & \bf 68.30$_{\pm 0.38}$ & \textbf{77.29}$_{\pm 0.24}$ / 45.80$_{\pm 0.10}$ / \textbf{55.41}$_{\pm 0.24}$ & 80 & \textbf{245} & \textbf{3.49} \\
    \quad w/o rethinking & 45.21$_{\pm 0.07}$ & 67.99$_{\pm 0.16}$ & 76.88$_{\pm 0.39}$ / 45.29$_{\pm 0.10}$ / 55.07$_{\pm 0.17}$ & {\bf 75} & 290 & 3.63 \\
    \hline
    \end{tabular}}
    }
    
    \subfloat[Results on WikiPerson]{
    \resizebox{\textwidth}{!}{
    \begin{tabular}{l l l l r r r}
    \hline
     \rowcolor[RGB]{237,237,237} \bf Models   & \bf BLEU & \bf ROUGE-L & \bf PARENT (P / R / F1) & \bf \#Param & \bf Latency$\downarrow$ & $\mathbf{I_{DEC}}$ $\downarrow$  \\
     \hline
    \multicolumn{7}{c}{\em One-Stage Systems} \\
    \hline

        \textproc{TableTransformer} & {\bf 25.11}$_{\pm 0.63}$ & {\bf 44.06}$_{\pm 0.56}$ & 61.13$_{\pm 0.89}$ / {\bf 52.08}$_{\pm 0.90}$ / {\bf 54.45}$_{\pm 0.41}$ & 92 & 2,092 & 62.42 \\
        \textproc{LevT} & 22.10$_{\pm 0.36}$ & 43.60$_{\pm 0.57}$ & {\bf 61.43}$_{\pm 0.44}$ / 49.58$_{\pm 0.00}$ / 53.65$_{\pm 0.16}$ & \bf 91 & \bf 449 & \bf 3.60 \\
    \hline
    \multicolumn{7}{c}{\em Two-Stage Systems} \\
    \hline

    \textproc{Content-Plan} & 25.17$_{\pm 0.78}$ & 44.47$_{\pm 0.03}$ & 62.09$_{\pm 0.55}$ / 53.63$_{\pm 0.44}$ / 56.68$_{\pm 0.29}$ & 187 & 2,708 & 82.61 \\
    \textproc{SANA} $\dag$ & \bf 25.23 & - & 65.69\ \quad\quad / 56.88\ \quad\quad / 59.96\ \quad\quad & 183 & - & - \\
    \quad w/o hard constraints $\dag$ & 24.97 & - & 64.72\ \quad\quad / 56.42\ \quad\quad / 59.29 & 183 & - & - \\
    \textproc{SANA} & 24.95$_{\pm 0.31}$ & \bf 45.35$_{\pm 0.13}$ & 69.26$_{\pm 0.83}$ / \textbf{58.16}$_{\pm 0.06}$ / 62.39$_{\pm 0.15}$ & 183 & 1,370 & 29.20 \\
    \quad w/o hard constraints & 22.37$_{\pm 0.38}$ & 45.08$_{\pm 0.15}$ & 69.10$_{\pm 0.49}$ / 56.62$_{\pm 0.38}$ / 61.38$_{\pm 0.28}$ & 183 & 1,217 & 28.92 \\
    \hdashline
    \textproc{Ours} & 25.11$_{\pm 0.35}$ & 45.23$_{\pm 0.31}$ & \textbf{69.72}$_{\pm 0.63}$ / 58.12$_{\pm 0.49}$ / \textbf{62.58}$_{\pm 0.36}$ & 97 & \bf 547 & \bf 4.83 \\
    \quad w/o rethinking & 24.45$_{\pm 0.28}$ & 44.87$_{\pm 0.21}$ & 68.17$_{\pm 0.74}$ / 57.45$_{\pm 0.52}$ / 61.55$_{\pm 0.43}$ & \bf 92 & 572 & 4.95 \\
    \hline
    \end{tabular}
    }
    }
    \caption{Results on WikiBio and WikiPerson test sets. Results marked with ``$\dag$'' are copied from previous studies while the other results are implemented in this work. Latency and $\mathrm{I_{DEC}}$ denote the average inference time and the average number of decoder iterations, respectively. Mean ($\pm$s.d.) over 4 seeds.}
    \label{tab:main_results}
\end{table*}

\subsection{Implementation Details}
Our method is implemented by {\tt fairseq} \citep{DBLP:conf/naacl/OttEBFGNGA19}. For fair comparison, all the involved systems use a similar configuration. Specifically, the vocabulary sizes on WikiBio and WikiPerson are $30$K and $50$K, respectively. The dimensions of token embedding, key embedding and position embedding are set to $420$, $80$, and $50$, respectively. All Transformer components used in our methods adopt the base Transformer setting with $d_{model} = 512$, $d_{hidden} = 2048$, and $n_{head} = 8$. The depth is $6$ for both the encoder and the decoder. 
Please refer to Appendix~\ref{sec:appendix_implement_training} for more details about training setting.
During inferance, the maximum iterations of the NAR model is 10 and 40 in WikiBio and WikiPerson, respectively.
We conduct experiments over 4 different random seeds and report the average scores.

\subsection{Main Results}
\label{sec:main_results}
Table~\ref{tab:main_results} shows the performance of our method and the baselines. 
For WikiBio, the NAR \textproc{LevT} model are approximately 3$\times$ faster than the AR \textproc{TableTransformer} model. 
However, the description quality of \textproc{LevT} is much lower than \textproc{TableTransformer}, regarding both fluency ($-$1.27 BLEU) and faithfulness ($-$4.62 PARENT-F1). Moreover, we observe that two-stage approaches can outperform the one-stage ones (e.g., \textproc{SANA} vs. \textproc{LevT}), indicating the superiority of explicitly content planning. However, they double the parameters scale and increase the inference latency. 
Surprisingly, our proposed method can combine the advantages of both the two kinds of baselines. 
On both WikiBio and WikiPerson, our approach can achieve comparable description quality with the strong two-stage baseline (i.e., \textproc{SANA}), while maintaining the model size and the inference speed. 
Compared with \textproc{SANA}, our model does not require external constraints to guarantee the appearance of planned tokens in the final output. 
The results also demonstrate the effectiveness of the newly proposed rethinking mechanism, confirming that the inter-dependency between different positions is essential for NAR-P, which can provide a better starting point for NAR-S.
Additionally, we notice that the latency increase without rethinking. We believe this is because removing this module reduces the content planning capability of the model, which in turn lowers the quality of the initial input to NAR-S, making the model require more iterations to satisfy the termination condition. 
The results on WikiPerson show a similar trend to WikiBio. An obvious difference is that the inference speed is much slower for all models, since the average description length is longer than WikiBio ($70.6$ vs. $26.1$). When generating longer sentences, the speedup of our method over the AR baseline is much higher. On both datasets, our method can achieve high description quality and inference efficiency at the same time.

\subsection{Analysis and Discussion}
\label{sec:anasysis_and_dis}
Due to the page limit, we have placed more experimental results and analyses in Appendix~\ref{sec:appendix_experimental_res}.
\paragraph{Content Planning}

As mentioned above, explicit content planning is important for table-to-text generation. We thus further investigate the content planning performance in Table~\ref{tab:plan_results}. We compare our method with two baselines: \textproc{PointerNetwork} and \textproc{NAR-P}. \textproc{PointerNetwork} is a widely used planning method for two-stage models~\citep{DBLP:conf/aaai/Puduppully0L19,DBLP:conf/acl/WangLYZZZY21}. The results indicate that our proposed \textproc{PTS} model performs comparable with \textproc{PointerNetwork}. \textproc{NAR-P} has a same architecture with \textproc{PTS}, the difference is that \textproc{NAR-P} is totally trained with the content planning task, while \textproc{PTS} is trained to perform both content planning and seaming. The results show that training the model with both content planning and seaming does not significantly affect the planning performance, which implies that learning the two tasks with a unified model does not decrease the model's planning ability.

\begin{table}[t]
    \centering
    \small
    \begin{tabular}{lcrr}
    \hline
    \rowcolor[RGB]{237,237,237} \bf Models  & \bf BLEU & \bf \#Param & \bf Latency \\
    \hline
    \multicolumn{4}{c}{\em WikiBio}\\
    \hline
    \textproc{PointerNetwork} &  64.97 & 76 & 261 \\
    \quad w/o beam search & 62.03 & 76 & 259 \\
    \hdashline
    \textproc{NAR-P} &  64.78 & 80 & 54 \\
    \quad w/o rethinking & 64.36 & 75 & 59 \\
    \hdashline
    \textproc{PTS-Plan} & 64.75 & 80  & 64 \\
    \quad w/o rethinking & 64.43 & 75 & 59 \\
    \hline
    \multicolumn{4}{c}{\em WikiPerson} \\
    \hline
     \textproc{PointerNetwork} & 52.42 & 91 & 851 \\
     \quad w/o beam search & 43.48 & 91 & 926 \\
     \hdashline
     \textproc{NAR-P} & 52.51 & 97 & 61 \\
     \quad w/o rethinking & 52.14 & 92 & 63 \\
    \hdashline
    \textproc{PTS-Plan} & 52.27 & 97 & 60 \\
    \quad w/o rethinking & 51.67 & 92 & 58 \\
    \hline
    \end{tabular}
    \caption{Performance of different content planning models. ``\textproc{NAR-P}'' is solely trained using the content planning objective, while ``\textproc{PTS-Plan}'' is to use the final \textproc{PTS} model to perform content planning.}
    \label{tab:plan_results}
\end{table}

\begin{figure}[t]
\centering
\includegraphics[width=0.9\columnwidth]{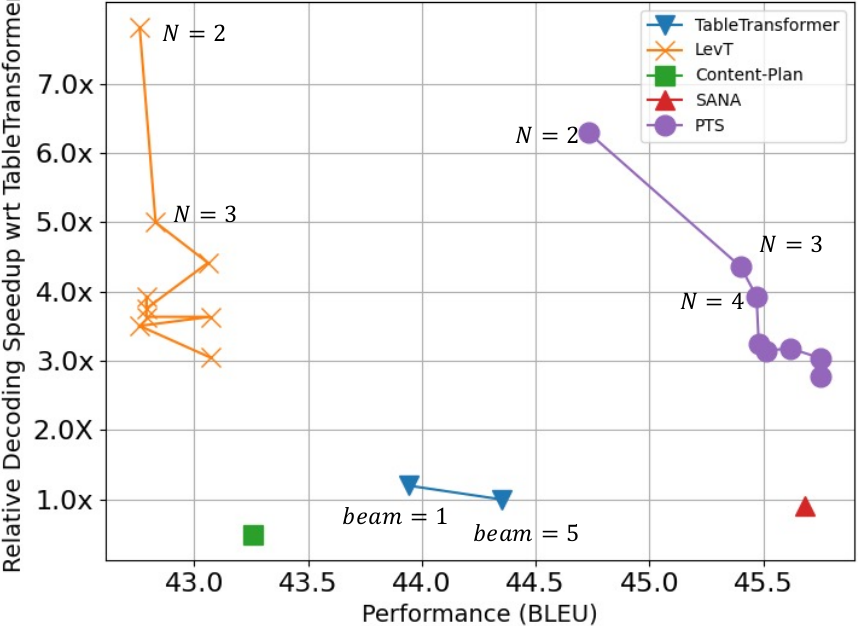}
\caption{Quality-Speed trade-off on the WikiBio test set. Quality is estimated by BLEU. For clarity, the inference speed is measured by the relative speedup with respect to \textproc{TableTransformer} with beam size = $5$. $N \in [1, 10]$ denotes the maximum iteration number.}
\label{fig:quality_efficiency_trade_off}
\end{figure}

\paragraph{Quality-Speed Trade-off}
Since \textproc{PTS} is an iterative NAR model, it is easy to balance the description quality and the inference speed by changing the number of the iterations.
As shown in Figure~\ref{fig:quality_efficiency_trade_off}, \textproc{PTS} achieves comparable performance with substantially higher speed up than the other involved models.
For \textproc{PTS}, increasing the iteration number can improve the description quality while reduce the inference speed. In practice, we can change the iteration number to meet different requirements under various application cases.

\begin{table}[]
    \centering
    \begin{tabular}{l r}
    \hline
    \bf Token & \bf Percentage (\%) \\
    \hline
        \texttt{,} & 8.01\%\\
        \texttt{.} & 7.69\% \\
        \texttt{-lrb-} & 7.28\% \\
        \texttt{-rrb-} & 7.26\% \\
        \texttt{a} & 5.21\% \\
        \texttt{is} & 5.07\% \\
        \texttt{born} & 4.51\% \\
        \texttt{the} & 3.12\% \\
        \texttt{an} & 2.74\% \\
        \texttt{was} & 2.27\% \\
        \texttt{and} & 2.41\% \\
        \texttt{in} & 2.26\% \\
        \texttt{--} & 2.13\% \\
        \texttt{of} & 2.12\% \\
        \texttt{who} & 1.64\% \\
    \hline
    \bf Total & 63.72\% \\
     \hline
    \end{tabular}
    \caption{Top 15 tokens generated by our proposed PTS at the seaming stage on WikiBio test set. \texttt{-lrb-} and \texttt{-rrb-} represent \texttt{(} and \texttt{)}, respectively.}
    \label{tab:top_15_words_generated_by_seaming}
\end{table}
\paragraph{Tokens Generated at the Seaming Stage}
To build a deeper understanding of the proposed PTS model, we investigate the problem of which tokens are most likely to be generated by PTS at the seaming stage on the WikiBio test set. Specifically, we first remove the words generated at the planning stage from the descriptions generated by PTS to obtain a word set. Then, we count the token frequency for the word in the set. We present the top 15 frequent tokens in Table~\ref{tab:top_15_words_generated_by_seaming}. As we can see, at the seaming stage, PTS is more likely to generate connective tokens, e.g., punctuation, and copulas, than the specific tokens existing in the input table, such as name, time, etc. And the connective tokens are mainly used to link planning words (seaming). The observation is consistent with our motivation and design that PTS first copies content from the input table to construct a plan sequence and then inserts tokens from a pre-defined vocabulary between plan tokens to generate a fluent description.
We believe this make our approach more interpretable and controllable.

\section{Human Evaluation}
To verify whether the system performance is consistent with what the automatic metrics show, we further conduct a human evaluation on the WikiBio test set. We randomly sample 50 instances from each model's generated outputs. Then, we invite three graduate students, whose English level is very high to understand the text, to score each generated text from 1 to 5 in terms of two criteria: Fluency (is the sentence fluency?) and Faithfulness (is the sentence related to the input table?). For each criterion, we average the scores from all annotators as the final score.
When evaluating, each annotator is provided the input tables as the references and does not know which model the generated text comes from. The results are summarized in Table~\ref{tab:human_evaluation}. As we can see, the overall trend of the human evaluation is similar to the automatic metrics in Table~\ref{tab:main_results}. First, the two-stage systems have an advantage over the one-stage ones in generating result fidelity. Second, our method is competitive to these table-to-text baselines regarding generation fluency and faithfulness. Meanwhile, our approach performs better than the \textproc{LevT} transferred from machine translation. These results indicate that introducing the content planning process in the NAR process and using its results as the initial decoder input can significantly improve the NAR model.
\begin{table}[t]
    \centering
    \small
    \begin{tabular}{llc}
    \hline
        \rowcolor[RGB]{237,237,237} \bf Models & \bf Fluency & \bf Faithfulness \\
         \hline
         \textproc{TableTransformer} & 3.24$_{\pm 0.77}$ & 2.89$_{\pm 0.86}$  \\
         \textproc{LevT} & 2.75$_{\pm 0.89}$ & 2.57$_{\pm 0.78}$  \\
         \textproc{Content-Plan} & 3.18$_{\pm 0.32}$ & 3.01$_{\pm 0.72}$  \\
         \textproc{SANA} & 3.31$_{\pm 0.74}$ & 3.26$_{\pm 0.69}$ \\
         \hline
         \textproc{Ours} & 3.42$_{\pm 0.58}$ & 3.42$_{\pm 0.59}$ \\
    \hline
    \end{tabular}
    \caption{Human evaluation results on WikiBio test set.}
    \label{tab:human_evaluation}
\end{table}
\section{Conclusion}
We propose a unified non-autoregressive framework, Plan-then-Seam (\textproc{PTS}), for table-to-text generation. Given a source table, \textproc{PTS} first generates the content plan in a fully NAR manner. Then we iteratively fill in the other surface tokens.
Experimental results demonstrate that \textproc{PTS} achieves a $3.0\sim5.6$ speedup with only $50\%$ model parameters compared with previous two-stage table-to-text models, without degrading the description quality.
Further analysis reveals that the success of \textproc{PTS} comes from the proposed NAR-P with a rethinking mechanism, whose content planning performance is comparable with AR models. By changeing the iteration number, \textproc{PTS} can balance the generation quality and inference efficiency for various practical application requirements.

\section*{Limitations}
As described in the paper, the content planning ability is important for table-to-text models. However, the planning performance of all the involved methods is still far from satisfactory. We will explore more advanced methods to improve the content planning performance. Moreover, we train all the models on WikiBio and WikiPerson from scratch, and the training cost is rather expensive: 2.5 days using 4 NVIDIA V100 32G GPUs. 
Lastly, this paper does not compare the pre-trained language models (PLMs)~\citep{devlin-etal-2019-bert, 2020t5,hui_2021_aaai_text2sql,Hui_2022_acl_text2sql}, though our approach may also benefit from some pre-trained table encoders, such as TAPAS~\citep{muller-etal-2021-tapas}.
The main reasons why we do not consider PLMs are that PLMs will bring an unfair comparison and bring more variables and may make our work lose focus. See Appendix~\ref{sec:appendix_experimental_setting} for detailed justification.
In the future, we will explore how pre-trained models, e.g., pre-trained table encoder TAPAS, can be used to improve our model's performance and accelerate the training process.
\section*{Ethics Statement}
We consider our work can make more researchers in table-to-text pay attention to the computational cost problem, which may benefit from saving the cost of the online table-to-text model. We experimented on the public datasets with no discriminatory or insulting sentences.
\bibliography{anthology,my_custom}
\bibliographystyle{acl_natbib}

\clearpage
\appendix

\section{More Experimental Results}
\label{sec:appendix_experimental_res}

\begin{table}[t]
    \centering
    \scalebox{0.75}{
    \begin{tabular}{lrrr}
    \hline
    \rowcolor[RGB]{237,237,237} \bf Models & \bf Token Repetitions (\%) & \bf Dist-1 & \bf Dist-2 \\
    \hline
    Gold & 12.58 & 5.6 & 22.83  \\
    \hline
    TableTransformer & 9.29 & 1.5 & 10.87 \\
    LevT & 14.59 & 2.1 & 13.97 \\
    \hline
    Content-Plan & 9.81 & 4.7 & 15.11 \\
    SANA & 9.93 & 5.0 & 19.32 \\
    \hline
    Ours & 8.22 & 4.9 & 18.15 \\
    \hline
    \end{tabular}
    }
    \caption{The token repetitions and diversity on WikiBio test dataset. Dist-1 and Dist-2 denote Distinct-1 and Distinct-2, respectively.}
    \label{tab:diversity}
\end{table}

\subsection{Token Repetitions and Diversity}
 Previous works manifest NAR model tends to predict the same token with high confidence, but at different positions, which is caused by the multi-modality problem. Therefore, we doubt whether the NAR table-to-text generation has any preference towards token repetitions and diversity. We measure the percentage of repetitive tokens in the generated sent as a proxy metric for the multi-modality problem \citep{gu2018non}. Additionally, we utilize Distinct-1 and Distinct-2 \citep{DBLP:conf/naacl/LiGBGD16} to evaluate the diversity of the output text. All results are summarized in Table \ref{tab:diversity}. We observe that the AR model \textproc{TableTransformer} significantly reduces the lexical diversity. Therefore, to better train the non-autoregressive model, AR model is usually used as a teacher model to reduce the complexity of the training corpus (Knowledge Distillation) \citep{gu2018non}. And then, \textproc{LevT} tends to generate repetitive tokens. We can see that, when explicitly modelling the content planning, two-stage methods can increase the tokens diversity. Especially, the content planning can substantially reduce the tokens repetitions for NAR models (e.g., \textproc{LevT} vs. SANA and \textproc{Ours}).

\subsection{Performance Bottleneck of Two-stage Model}
\begin{table}[t]
    \centering
    \small
    \begin{tabular}{lcc}
    \hline
        \rowcolor[RGB]{237,237,237} \bf Models & \bf BLEU & \bf PARENT(P / R / F1) \\
         \hline
         \textproc{Content-Plan} & 43.72 & 76.55 / 38.79 / 49.02 \\
         \quad + Golden Plan &  51.16 & 76.32 / 47.33 / 56.45 \\
         \hdashline
         \textproc{SANA} & 45.68 & 76.79 / 45.64 / 55.12 \\
         \quad + Golden Plan &  54.30 &  80.03 / 51.02 / 61.01 \\
         \hdashline
         \textproc{Ours} & 45.75 &  77.34 / 45.91 / 55.48 \\
         \quad + Golden Plan & 55.50 &  79.59 / 51.92 / 61.14 \\
    \hline
    \end{tabular}
    \caption{Effect of golden plan on two-stage methods.}
    \label{tab:two_stage_with_gold_plan}
\end{table}

We provide the ground-truth content plan to the models at the second stage, and the results are summarized in Table \ref{tab:two_stage_with_gold_plan}. When fed with the golden plan, all the two-stage models achieves better fluency and faithfulness. The results indicate that the quality of content planning is a important  bottleneck for two-stage table-to-text approaches.
\subsection{Case Study}
Table \ref{tab:case_study} shows the descriptions generate by \textproc{PTS} from the test set of WikiBio. First, we observe that when the number of tokens in the generated plan is relatively small, the $1$-st pointer predictor can generate a precise content plan. However, when the number of planning tokens increases, it tends to produce repetitive and incorrect ones. We consider this is because the fully NAR generation cannot accurately model the dependencies between planning tokens. After introducing the \textit{rethinking} mechanism, the $2$-nd pointer predictor can determine if the token is incorrect or repeated with others and calibrate it by looking at the tokens in other positions. Therefore, the model can generate a more precise plan.

\begin{table*}[t]
    \centering
        \scalebox{1.0}{
    \begin{tabular}{c|l|l|l}
    \hline
    \rowcolor[RGB]{237,237,237} \multicolumn{4}{c}{First Example} \\
    \hline
       \bf Source Table  & \multicolumn{3}{l}{\parbox[c][1.5cm]{1.5\columnwidth}{<\textbf{Name}: sean macias>, <\textbf{Birth Place}: california>, <\textbf{Known For}: litigation>, <\textbf{Occupation}: lawyer>, <\textbf{Nationality}: american>, <\textbf{Article Title}: sean macias>.}} \\
    \hline
       \bf Flatten Table & \multicolumn{3}{l}{\parbox[c][2cm]{1.5\columnwidth}{
       (\textbf{Name}, sean, 1, 2), (\textbf{Name}, macias, 2, 1), (\textbf{Birth Place}, california, 1, 1), (\textbf{Known For}, litigation, 1, 1), (\textbf{Occupation}, lawyer, 1, 1), (\textbf{Nationality}, american, 1, 1), (\textbf{Article Title}, sean, 1, 2), (\textbf{Article Title}, macias, 2, 1).
       }}\\
      \hline
      \bf Rerence & \multicolumn{3}{l}{\parbox[c][1cm]{1.5\columnwidth}{sean ernesto macias -lrb- born 31 october 1972 -rrb- is a pasadena-based litigation lawyer known for handling high-profile cases .}} \\
     \hline
        \multirow{5}{*}{\bf \textproc{PTS}} & \multirow{2}{*}{1st iteration} & 1st pointer predictor & sean macias american litigation lawyer  \\
        & & 2nd pointer predictor & sean macias american litigation lawyer  \\
        \cline{3-4}
        & 2nd Iteration & \multicolumn{2}{l}{sean macias \textcolor{green}{is an} american litigation lawyer \textcolor{green}{.}} \\
        \hline
    \rowcolor[RGB]{237,237,237} \multicolumn{4}{c}{Second Example} \\
    \hline
       \bf Source Table  & \multicolumn{3}{l}{\parbox[c][3cm]{1.5\columnwidth}{<\textbf{Name}: dave green>, <\textbf{Poisition}: punter placekicker >, <\textbf{Number}: 4>, <\textbf{Birth Date}: 21 september 1949>, <\textbf{Debutyear}: 1973>, <\textbf{Finalyear}: 1978>, <\textbf{Draftyear}: 1972>, <\textbf{Draftround}: 17>, <\textbf{Draftpick}: 418>, <\textbf{College}: ohio university>, <\textbf{Statlabel}: punts punting yards punting avg 446 17,883 40.1>, <\textbf{Nfl}: re162282>, <\textbf{Brith Place}: mason city iowa> ,<\textbf{Article Title}: dave green -lrb- american football -rrb->.}} \\
    \hline
       \bf Flatten Table & \multicolumn{3}{l}{\parbox[c][4cm]{1.5\columnwidth}{
       (\textbf{Name}, dave, 1, 2), (\textbf{Name}, green, 2, 1), (\textbf{Poisition}, punter, 1, 2), (\textbf{Poisition}, placekicker, 2, 1), (\textbf{Number}, 4, 1, 1), (\textbf{Birth Date}, 21, 1, 3), (\textbf{Birth Date}, september, 2, 2), (\textbf{Birth Date}, 1949, 3, 1), (\textbf{Debutyear}, 1973, 1, 1), (\textbf{Finalyear}, 1978, 1, 1), (\textbf{Draftyear}, 1972, 1, 1), ..., (\textbf{Birth Place}, mason, 1, 3), (\textbf{Birth Place}, city, 2, 2), (\textbf{Birth Place}, iowa, 3, 1), (\textbf{Article Title}, dave, 1, 6), (\textbf{Article Title}, green, 2, 5), (\textbf{Article Title}, -lrb-, 3, 4), (\textbf{Article Title}, american, 4, 3), (\textbf{Article Title}, football, 5, 2), (\textbf{Article Title}, -rrb-, 6, 1).
       }}\\
      \hline
      \bf Rerence & \multicolumn{3}{l}{\parbox[c][1cm]{1.5\columnwidth}{dave green -lrb- born september 21 , 1949 in mason city , iowa -rrb- is a former punter and placekicker in the national football league .}} \\
     \hline
        \multirow{8}{*}{\bf \textproc{PTS}} & \multirow{2}{*}{1st iteration} &  1st pointer predictor & \parbox[c][1.5cm]{0.8\columnwidth}{dave \textcolor{red}{dave} september 21 1949 \textcolor{red}{american} city iowa american football punter placekicker football}  \\
        & & 2nd pointer predictor & \parbox[c][1.5cm]{0.8\columnwidth}{dave green september 21 1949 mason city iowa american football punter placekicker football}  \\
        \cline{2-4}
        & 2nd Iteration & \multicolumn{2}{l}{\parbox[c][2cm]{1.2\columnwidth}{dave alan green -lrb- born september 21 , 1949 in mason city , iowa -rrb- is a former american football punter and placekicker in the national football league .}} \\
        \cline{2-4}
        & 3rd Iteration & \multicolumn{2}{l}{\parbox[c][2cm]{1.2\columnwidth}{dave green \textcolor{green}{-lrb- born} september 21 \textcolor{green}{,} 1949 \textcolor{green}{in} mason city \textcolor{green}{,} iowa \textcolor{green}{-rrb- is a former} american football punter \textcolor{green}{and} placekicker \textcolor{green}{in the national} football \textcolor{green}{league .}}} \\
        \hline
    \end{tabular}
    }
    \caption{Two examples from the WikiBio test set that illustrates how PTS generates a description for a source table by planning and then seaming. The incorrect and repetitive planning tokens are in red.}
    \label{tab:case_study}
\end{table*}

\section{More Implementation Details}
\subsection{Training and Hyper-parameter Settings}
\label{sec:appendix_implement_training}
All models are optimized by Adam~\citep{Kingma:2015:Adam}. We use the same learning rate schedule as presented in \citet{DBLP:conf/nips/VaswaniSPUJGKP17}. The maximum value of the learning rate is 5e-4 and the warmup step is set to 10K. 
The maximum training step is set to 300K. We use the validation BLEU for early stopping and explore $\lambda=[0.05, 0.08]$. 
During inferance, we use beam search with a beam size 5 for the autoregressive models and the maxinum decoding lengths are set to 80 and 160 in WikiBio and Wikiperson. For non-autoregressive models, we set the maximum iterations as 10 and 40 in WikiBio and WikiPerson, respectively.

\subsection{Experimental Setting Details}
\label{sec:appendix_experimental_setting}
To rule out the effect of model architecture on the inference speed and make a fair comparison, we only compare our method to some table-to-text models built on the Transformer model. For the one-stage models, we chose the autoregressive TableTrasnforme and non-autoregressive LevT. For the two-stage methods, we compare with Content-Plan and SANA. 
Both planning generation and tableau generation of the former are autoregressive, while the second stage of the latter is a non-autoregressive process. All these baselines employ the same table encoder as ours.
Additionally, the original Content-Plan is implemented by LSTM. To make a fair comparison, we re-implemented Content-Plan by replacing its LSTM-based encoder and decoder with Transformer-based ones. And the transformer setting is the same as our model.

We do not consider the pre-trained models (PLMs) in this paper, though our model's performance may be significantly improved by initializing our table encoder with the pre-trained one, such as TAPAS \citep{herzig-etal-2020-tapas}. The reasons why we do not consider PLMs are as follows: 
\begin{itemize}

    \item We consider PLMs will bring an unfair comparison. Because most PLMs (e.g., TAPAS, T5~\citep{2020t5}) are pre-trained on Wikipedia data, and WikiBio and WikiPerson are built from Wikipedia. It may lead to data leakage. Moreover, to our best knowledge, most of the works in the NAR machine translation (please refer to Appendix~\ref{sec:appendix_nar_translation})  do not compare with PLMs. 
    \item This paper focuses on comparing inference speed and quality under a similar model architecture rather than improving the model performance. And our experimental setting is fair, and all the baselines employ a similar setting as our model. Additionally, in related domains such as neural machine translation, previous work~\citep{zhu2020incorporating} indicates that simply initializing the encoders of sequence-to-sequence models with the pre-trained BERT \citep{devlin-etal-2019-bert} will actually hurt the performance. And directly fine-tuning NAR sequence-to-sequence models initialized by BERT is very unstable and sensitive to the learning rate~\citep{guo2020incorporating}. Therefore, though pre-trained checkpoint may benefit our model, it will bring more variables and may make our work lose focus. We leave this for feature work.
    
\end{itemize}

\subsection{Content Plan Annotation}
We follow previous work \citep{DBLP:conf/acl/WangLYZZZY21} to employ the heuristic method to obtain the content plan annotation for WikiBio and WikiPerson. Specifically, we start by counting the tokens that appear both in the table and in the corresponding description. Then we remove the stop tokens in the tokens collection and sort the rest of the tokens by the their positions in the description in ascending order. The sorted sequence is regard the content planning sequence. We refer the readers to \citet{DBLP:conf/acl/WangLYZZZY21}'s paper for more details.

\section{Non-autoregressive Neural Machine Translation}
\label{sec:appendix_nar_translation}
Recently, autoregressive (AR) models have achieved outstanding performances in natural language generation tasks \citep{2020t5}. However, AR is quite time-consuming when generating target sentences, especially for long sentences. To overcome this problem and accelerate decoding, \citet{gu2018non} first propose the non-autoregressive generation (NAR) for machine translation, which generates all the target tokens in parallel and hugely increases the inference speed. Therefore, much attention has been attracted to NAR with impressive progress \citep{DBLP:conf/icml/SternCKU19,DBLP:conf/acl/QianZBWQ0YL20,song2021alignart,qian2021glancing}. However, compared with AR models, the generation quality is sacrificed because NAR breaks the conditional dependence assumption that prevents a model from properly capturing the highly multi-modal distribution of target translations, which is called the "multi-modality" problem \citet{gu2018non}. 
To mitigate the problem, some studies \citep{DBLP:conf/emnlp/LeeMC18,DBLP:conf/icml/SternCKU19,DBLP:conf/emnlp/GhazvininejadLL19,DBLP:conf/nips/GuWZ19,DBLP:conf/emnlp/SahariaCSN20} propose the iterative NAR models which need $N$ iterations for inference and keep the non-autoregressive property in every iteration step. More specifically, the generated results of the previous iteration will be fed into the decoder again for refinements. In this way, partial target information is provided in each iteration step.

\end{document}